\documentclass[runningheads]{llncs}
\usepackage{amsmath}
\usepackage[T1]{fontenc}
\usepackage{hyperref}
\usepackage{xcolor}

\usepackage{graphicx}
\begin{document}
\title{ADASSM: Adversarial Data Augmentation in Statistical Shape Models From Images}
\titlerunning{ADASSM: Adversarial Data Augmentation in SSM from Images}
\author{Mokshagna Sai Teja Karanam$^{1,2}$ \and Tushar Kataria$^{1,2}$ \and Krithika Iyer$^{1,2}$ \and Shireen Y. Elhabian$^{1,2,3}$}

\authorrunning{M. Karanam et al.}
\institute{Kahlert School of Computing, University Of Utah \and
Scientific Computing and Imaging Institute, University of Utah \and Corresponding Author\\ \{mkaranam@sci, tushar.kataria@sci, krithika.iyer@, shireen@sci\}.utah.edu}

\maketitle
\begin{abstract}
Statistical shape models (SSM) have been well-established as an excellent tool for identifying variations in the morphology of anatomy across the underlying population. Shape models use consistent shape representation across all the samples in a given cohort, which helps to compare shapes and identify the variations that can detect pathologies and help in formulating treatment plans. In medical imaging, computing these shape representations from CT/MRI scans requires time-intensive preprocessing operations, including but not limited to anatomy segmentation annotations, registration, and texture denoising. Deep learning models have demonstrated exceptional capabilities in learning shape representations directly from volumetric images, giving rise to highly effective and efficient Image-to-SSM networks. Nevertheless, these models are data-hungry and due to the limited availability of medical data, deep learning models tend to overfit. Offline data augmentation techniques, that use kernel density estimation based (KDE) methods for generating shape-augmented samples, have successfully aided Image-to-SSM networks in achieving comparable accuracy to traditional SSM methods. However, these augmentation methods focus on shape augmentation, whereas deep learning models exhibit image-based texture bias resulting in sub-optimal models. This paper introduces a novel strategy for on-the-fly data augmentation for the Image-to-SSM framework by leveraging data-dependent noise generation or texture augmentation. The proposed framework is trained as an adversary to the Image-to-SSM network, augmenting diverse and challenging noisy samples. Our approach achieves improved accuracy by encouraging the model to focus on the underlying geometry rather than relying solely on pixel values.

\keywords{Statistical Shape Model  \and Data Augmentation \and Adversarial Training}
\end{abstract}

\section{Introduction}

\hspace{0.5cm}Statistical shape modeling (SSM) is widely used in the fields of medical image analysis and biological sciences for studying anatomical structures and conducting morphological analysis. It enables shape analysis by facilitating the understanding of the geometrical properties of shapes that are statistically consistent across a population. SSM has diverse applications in neuroscience \cite{zhao2008hippocampus,gerig2001shape}, cardiology \cite{bhalodia2018deep}, orthopedics \cite{harris2013statistical,bhalodia2020quantifying}, and radiology \cite{bharath2018radiologic,gardner2013point}.

Optimization-based SSM \cite{cates2017shapeworks} methods typically involve anatomy segmentation, data preprocessing (e.g., image resampling, denoising, rigid registration), and optimizing population-level shape representation i.e., correspondence points (or particles), all of which require substantial expertise-driven workflow, involving intensive preprocessing that can be time-consuming. Deep learning approaches for SSM, train networks to learn the functional mapping from unsegmented images to statistical representations of anatomical structures \cite{bhalodia2018deep,bhalodia2021deepssm,adams2020uncertain,adams2023fully,adams2022images,tothova2018uncertainty}. This shift towards deep learning-based methods offers a more efficient and automated approach to SSM, bypassing the need for extensive manual preprocessing and leveraging the power of neural networks to learn directly from raw imagery data \cite{adams2023fully,xu2023image2ssm}. However, deep learning models are notorious for requiring enormous quantities of data to achieve acceptable performance \cite{krizhevsky2017imagenet}, necessitating the use of data augmentation to supplement the available training data \cite{bhalodia2018deep}.

In the field of deep learning for medical image analysis, data augmentation plays a crucial role \cite{abdollahi2020data,chlap2021review,hussain2017differential}. Nevertheless, unlike computer vision applications, acquiring a substantial number of segmented medical images is difficult due to privacy concerns, the substantial human effort and expertise required, and the intensive preprocessing involved \cite{gao2021enabling}. Off-the-shelf data augmentation methods may not generate augmented samples that promote invariances and improve the task-specific generalizability of the model \cite{geiping2022much}. Therefore, having a large amount of data supplemented with challenging task-specific variations would be extremely beneficial for training deep neural networks and improving model performance. In the field of medical imaging, attempts have been made to employ task-driven automatic data augmentation techniques \cite{cubuk2018autoaugment} for image segmentation and classification \cite{xu2020automatic,gao2021enabling,chlap2021review}. For regression tasks, various strategies have been proposed for handling data augmentation that includes, data-dependent shape augmentation for Image-to-SSM networks \cite{bhalodia2018deepssm,bhalodia2021deepssm,adams2020uncertain} and a mix-up \cite{zhang2017mixup} based augmentation by interpolating input samples using the similarity of labels \cite{yao2022c}. 

DeepSSM \cite{bhalodia2021deepssm,bhalodia2018deepssm}, and other variants \cite{adams2020uncertain,adams2022images,adams2023fully}, learn to map unsegmented images to shape models, exhibiting comparable performance to traditional SSM \cite{cates2017shapeworks} methods, as well as in downstream tasks such as atrial fibrillation recurrence \cite{bhalodia2018deep,harris2013statistical}. DeepSSM relies heavily on offline shape-based data augmentation via kernel density estimation (KDE) in the linear principal component analysis (PCA) subspace \cite{bhalodia2018deepssm,bhalodia2021deepssm}. The DeepSSM shape augmentation approach entails using generative modeling to sample shapes from probability distribution estimated via KDE. Existing offline methods have three deficiencies: (1) the generation of augmented samples is independent of the task (shape modeling) at hand, (2) the augmentation process focuses on shape augmentation rather than incorporating noise/texture augmentation, neglecting the inherent texture bias often present in deep learning models \cite{hermann2020origins}, which can lead to sub-optimal models in shape analysis, and (3) they require extensive offline data compilation, which is time intensive and resource consuming.

We draw inspiration from adversarial domain adaptation \cite{ganin2016domain} and adversarial data augmentation for classification tasks \cite{gao2021enabling} and adapted these ideas to regression tasks with application to Image-to-SSM networks. The regression task poses a greater challenge as it is not straightforward to generate challenging adversarial samples for the learning task at hand due to the absence of label-separating hyperplanes. Consequently, we focus our methodology on generating noise-augmented samples as an alternative to KDE based shape augmentation \cite{bhalodia2018deepssm,bhalodia2021deepssm}. The proposed method implicitly drives the model to attend to the shape of the underlying object of interest instead of explicit shape augmentation \cite{bhalodia2018deepssm,bhalodia2021deepssm}. We also demonstrate that data- and task-dependent noise augmentation is better than off-the-shelf noise augmentation with varied variance levels. The proposed augmentation approach is generic enough to be used for any Image-to-SSM network, but here we focus on DeepSSM \cite{bhalodia2018deepssm} to showcase the efficacy of the on-the-fly noise augmentation vs offline shape and noise augmentation.

As we focus on image noise augmentation for this work, the shape representation of the augmented images should not be affected. As a result, we employ a contrastive loss to inform the deep learning model that noisy and their corresponding original images should be projected to the same latent representation. This contrastive loss acts as a regularizer to both the augmentation framework and the Image-to-SSM network.

\noindent The contributions of this paper are as follows:-

\begin{itemize}
    \item A computationally efficient, automated, on-the-fly adversarial data augmentation method for regression tasks with better generalization.
    \item A contrastive loss based regularization that enables enhanced noise generation that is more task- and data-dependent.
    \item Extensive experiments with Image-to-SSM and downstream tasks on left atrium and femur datasets show the efficacy of the proposed approach.
\end{itemize}

\section{Methodology}
\begin{figure}[!htb]
    \centering
    \includegraphics[scale=0.37]{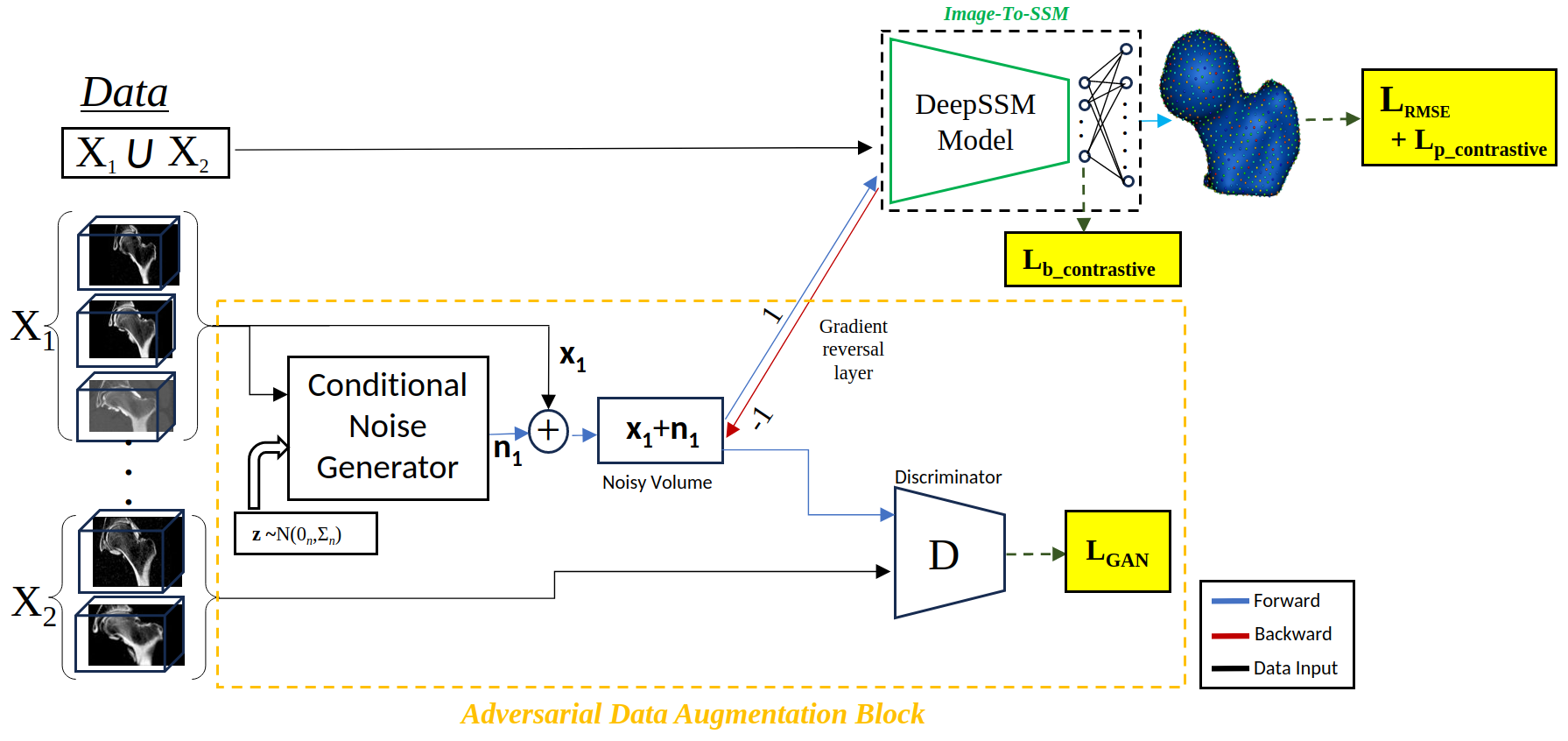}
    \caption{\textbf{ADASSM Block Diagram.} \textbf{Conditional Noise Generator} uses the input volumes ($\textbf{x}_1$) to generate noise ($\textbf{n}_1$), which is added back to the input volume. \textbf{Discriminator} used the remaining volumes ($\textbf{X}_2$) for GAN loss($\mathbf{L}_{GAN}$ ) to ensure the generated data is in distribution. The noise generator and DeepSSM are connected by a \textbf{gradient reversal layer}, which sets up the second adversarial training paradigm. Both the noisy and original volumes are used to train the Image-to-SSM framework (DeepSSM). $\mathbf{L}_{RMSE}$ is the correspondences RMSE loss, $\mathbf{L}_{p\_contrastive}$ \& $\mathbf{L}_{b\_contrastive}$ are correspondence and bottleneck contrastive loss.}
    \label{fig:method:block_diagram}
\end{figure}

This section explains the details of the proposed method and the regularization losses. The block diagram for the proposed approach is shown in Figure~\ref{fig:method:block_diagram}.
\subsection{Adversarial Data Augmentation Block}
The proposed framework for augmentation aims to enhance the performance of the Image-to-SSM network by generating data-dependent noise. This architecture can be especially useful in the context of SSM tasks, where the input data size is typically limited. Due to the paucity of data samples, deep learning models may suffer from overfitting, leading to poor generalization performance on unseen data. To address this, we integrate an adversarial \cite{ganin2016domain,gao2021enabling} data augmentation approach for regression tasks with applications in shape modeling. The generator produces adversarial data samples that are difficult for the shape model to project into statistical shape representation. 

\indent \textbf{Conditional Noise Generator:} Conditional noise generator receives an image $\mathbf{x_1}$ from input set $X_1$ and a randomly sampled Gaussian vector, $\mathbf{z}$ as inputs. It then generates noise vector $\mathbf{n_1}$, which is subsequently added to the original input volume $\mathbf{\hat{x}_1} =\mathbf{x_1+n_1}$, resulting in the generation of a noisy augmented sample. 

    \begin{equation}
        \mathbf{\hat{x}_1} = \mathcal{R} * G(\mathbf{z},\mathbf{x}_1) \oplus \mathbf{x}_1
    \end{equation}
Here, $\mathcal{R}$  is a hyperparameter for noise perturbation range, G denotes the conditional generator, and $\oplus$ represents voxelwise addition. To achieve controlled augmentations and minimize excessive perturbations, a regularization loss based on total variation (TV) is incorporated \cite{gao2021enabling}. TV loss also enables the generation of noise variations that exhibit smooth transitions, which is crucial for capturing the inherent features of real-world images.
\begin{equation}
        \mathbf{L}_{TV} = ||G(\mathbf{z},\mathbf{x}_1)||_2
\end{equation}
\indent \textbf{Discriminator:} To further regularize the noise generation a discriminator is used. This discriminator uses the remaining input samples from set $\textbf{X}_2$, as reference distribution, and noisy samples are treated as samples of input distributions. The objective of the discriminator is to assist the generator in producing realistic noise while ensuring that the noisy augmented sample comes from the same distribution as the original data. The generative adversarial network (GAN) aims to strike a balance between meaningful data-specific augmentations and excessive perturbations. The loss of the block is given in the equation below:-
\begin{equation}
\mathbf{L}_{\text{GAN}} = \underset{G}{\text{min}} \; \underset{D}{\text{max}} \; \mathbf{E}_{X\in \Omega} \; [\log D(\mathbf{x_2})] + \mathbf{E}_{X\in \Omega} [\log(1-D(G(\mathbf{z},\mathbf{x_1}) \oplus \mathbf{x_1}))]\hspace{0.2cm} +  \beta \mathbf{L}_{TV}
\end{equation}
where $\beta$ is hyperparameter.

\subsection{Adversary to Image-To-SSM network}
The noisy augmented sample is fed into the Image-To-SSM network to obtain the predicted shape representation($\mathbf{\hat{y}}$), which is compared to the original shape representation via an RMSE loss. 
\begin{equation}
     \mathbf{L}(\mathbf{\hat{y}},\mathbf{y}) = RMSE(\mathbf{y},\text{DeepSSM}(\mathbf{\hat{x}_1}))
\end{equation}

The Image-to-SSM network and GAN framework are put in an adversarial relationship with the help of a gradient reversal layer, as shown in the block diagram in Figure~\ref{fig:method:block_diagram}. The objective function in equation~\ref{rmse} aims to minimize the Image-To-SSM network error while maximizing the conditional generator G perturbations, setting up a second adversarial objective:
\begin{equation}\label{rmse}
    \mathbf{L}_{\text{RMSE}} = \mathbf{E}_{X,Y\in \Omega} [\underset{M}{\text{min}} \; \underset{G}{\text{max}} \; \mathbf{L}(\mathbf{\hat{y}}, \mathbf{y})]
\end{equation}
The framework above allows the augmentation model to search along the adversarial direction, leading to the generation of challenging noise augmentations that facilitate the learning of more robust shape features.

\subsection{Image-To-SSM Network:} The DeepSSM Model employs a deterministic encoder and a deterministic linear decoder. The reconstructed correspondences are obtained as the output of the Image-to-SSM network, providing both shape representation as well as low dimensional latent features for each input volume.

\subsection{Shape Regularization Loss:} Noise augmentation affects only the texture of the input volume and does not affect the underlying shape. We hypothesize that the shape representation (after DeepSSM) of the augmented noisy volume ($\mathbf{\hat{x}_1}$) and original volume ($\mathbf{x}_1$), should be closer in the shape space. We use a contrastive loss in equation~\ref{contrastive_loss} as an additional regularizer to ensure that both the Image-to-SSM network and GAN account for this.
\begin{equation}\label{contrastive_loss}
    \mathbf{L}_{contrastive} = -\log(\frac{\exp{(sim(DeepSSM(\mathbf{x_1}),DeepSSM(\mathbf{\hat{x}_1})))}}{\sum^N \exp(sim(DeepSSM(\mathbf{x_1}),DeepSSM(\mathbf{\hat{x}_1})))})
\end{equation}

We propose to use contrastive loss at two different latent representations as shown in Figure~\ref{fig:method:block_diagram}: 
\begin{enumerate}
    \item\textit{Correspondences} ($\mathbf{L}_{\text{p\_contrastive}}$, PC), which are the predicted SSM representation by the Image-to-SSM network. This loss ensures that the augmentation does not effect the final statistical shape representation of the augmented image.
    \item\textit{Bottleneck} ($\mathbf{L}_{\text{b\_contrastive}}$, BC), which is the low-dimensional space representation obtained from the Image-to-SSM network. This loss helps the model learn the same latent representation despite noise augmentation. 
\end{enumerate}

These regularization losses encourage both the generator and shape model to focus more on shape-related information during the learning process and factor out texture variations. The overall objective function for the proposed model is:
\begin{equation}
    \mathbf{L} = \alpha \mathbf{L}_{GAN} + \mathbf{L}_{RMSE} + \lambda_1 \mathbf{L}_{\text{b\_contrastive}} + \lambda_2 \mathbf{L}_{\text{p\_contrastive}}
\end{equation}
where $\alpha,\lambda_1,\lambda_2$ are hyperparameters. 

\section{Results}
We use the same Image-to-SSM (DeepSSM \cite{bhalodia2021deepssm}) model architecture across all experiments to ensure that variations in model performance can only be attributed to different augmentation techniques. As a baseline for comparison, we train an Image-to-SSM architecture without any augmentations (\textit{NoAug}). Additionally, we train another model using KDE \cite{bhalodia2021deepssm} augmentation (\textit{KDE \cite{bhalodia2021deepssm}}). We also compute two other baselines with off-the-shelf Gaussian noise augmentation with different variances($\sigma=1 \text{ and } 10$). 
\subsection{Metrics}
    \hspace{0.5cm} \textbf{Root Mean Squared Error (RMSE):}{ To measure the error, we calculate the average relative mean squared error (RMSE) between the predicted 3D correspondences and this is achieved by computing the RMSE for the x, y, and z coordinates and averaging them as shown in (7) }
\begin{equation}
    RMSE = \frac{1}{3} ( RMSE_x + RMSE_y + RMSE_z)
\end{equation}
For N 3D correspondences, $RMSE_x = \sqrt{\frac{||C_x - C_x^{'}||_2^2}{N}}$. The same calculation is applied to $RMSE_y$ and $RMSE_z$ for the respective coordinates. Additionally, we calculate the RMSE error for each correspondence point as, $RMSE_i = \sqrt{\frac{||C_x^i - C_x^{'i}||_2^2 + ||C_y^i - C_y^{'i}||_2^2 + ||C_z^i - C_z^{'i}||_2^2}{3}}$

The per-point RMSE helps us assess the accuracy of DeepSSM in modeling various local anatomical features.  For all experiments, the shape representations were calculated on same test data (held-out data) using the trained DeepSSM model and were only used for inference. 
\\

    \textbf{Surface-to-Surface Distance (mm):} {The surface-to-surface distance is measured between the ground truth mesh and the mesh reconstructed from the predicted correspondences by DeepSSM. This distance provides a more precise measure of how well the correspondences adhere to the shape and indicates their suitability for anatomy segmentation. 
    
Furthermore, we validate the effectiveness of DeepSSM with the learned shape representations by utilizing its correspondences for various downstream analysis applications. The specific downstream applications vary for each dataset and are described in separate subsections below.}

\subsection{Femur}

\hspace{0.15in} \textbf{Data Description and Processing:}
The femur dataset consists of 49 CT images of the femur bone, of which 42 are considered healthy with no morphological abnormalities. DeepSSM \cite{bhalodia2018deepssm,bhalodia2021deepssm} requires generating point distribution models (or correspondences) for the training images. We use ShapeWorks \cite{cates2017shapeworks} to optimize a shape model with 1024 correspondences. Along with the training and validation data, we also randomly selected 7 controls and 2 CAM-FAI scans for testing the DeepSSM Model. To meet GPU memory requirements, each image is downsampled by a factor of 2 from $260 \times 184 \times 235$ (0.5mm isotropic voxel spacing) to isotropic voxel spacing of 1mm with dimensions of $130 \times 92 \times 117$. The training images are divided into training and validation sets with 80\% -20\% split.

\textbf{Training Specifics: }Empirically we set $\alpha = 1$ and $\beta = 0.1$. The default parameters and configuration are used to get the result for KDE \cite{bhalodia2021deepssm} augmentation for femur dataset. The training process involves optimizing the loss on correspondences for 1500 epochs, employing a data augmentation framework based on validation loss. For all ADASSM experiments, involving the proposed regularization losses, a learning rate of 5e-5 is utilized while ADASSM itself is trained with a learning rate of 1e-5. The generator and discriminator learning rates are set to 5e-3, except for ADASSM+BC+PC, which employs a learning rate of 1e-3. A batch size of 4 is used for training all the proposed models and baseline models. In the generator, $\mathcal{R}$ is set to 500 for ADASSM experiments.

\begin{figure}[!htb]
    \centering
    \includegraphics[scale=0.39]{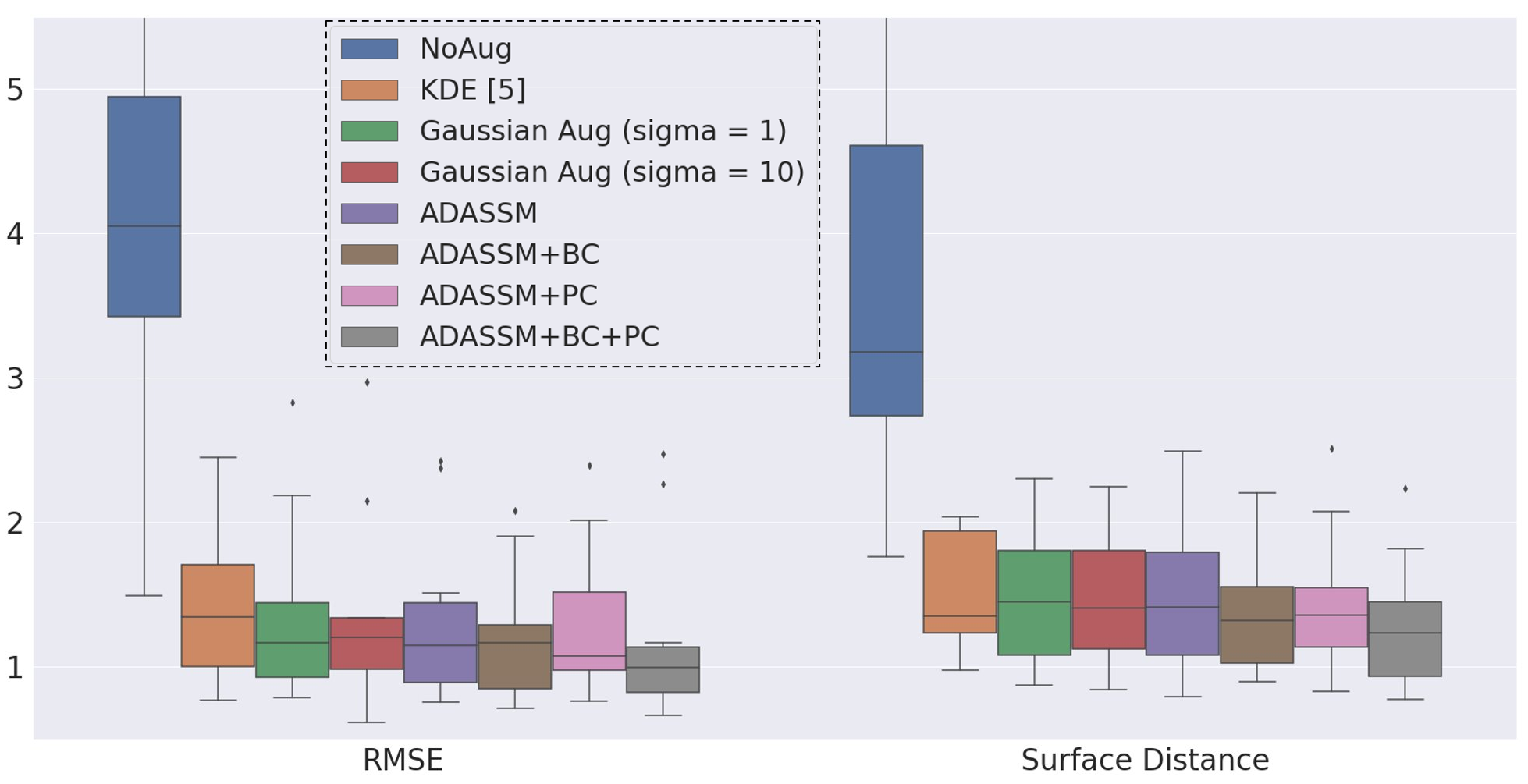}
    \caption{\textbf{Femur Test Results.} The \textit{RMSE} is the Euclidean distance between the ground truth and predicted correspondences for all the test samples. The \textit{}{Surface-to-Surface distance (mm)} is computed by comparing the reconstructed mesh using the ground truth correspondences and the predicted correspondences for all the test samples. $\lambda_1$ for ADASSM+BC is 0.5, $\lambda_2$ for ADASSM+PC is 0.1 and $\lambda_1,\lambda_2$ for ADASSM+BC+PC is 0.5. Y-axis is the magnitude of the errors displayed.} 
    \label{fig:result:femur}
\end{figure}

\textbf{Evaluation and Analysis:} Figure~\ref{fig:result:femur} visualizes the RMSE results alongside the surface-to-surface distance.Augmenting the DeepSSM model with Gaussian noise with different variances without any adversarial training improves the RMSE when compared to the KDE augmentation\cite{bhalodia2021deepssm}. However, an increase in the surface distance of the predicted correspondences indicates misalignment with the ground truth femur bone segmentation. This result shows that standard noise augmentation can provide better results for Image-to-SSM networks.

We can observe that for the proposed ADASSM (data-dependent noise augmentation), RMSE results are better compared to both KDE \cite{bhalodia2021deepssm} and Gaussian noise. The addition of contrastive regularization losses further improves RMSE error and surface distance, which proves that data-dependent noise-augmented samples are better compared to Gaussian augmented samples and shape augmentation \cite{bhalodia2021deepssm}.  

\begin{figure}[!htb]
     \centering
    \includegraphics[scale=0.63]{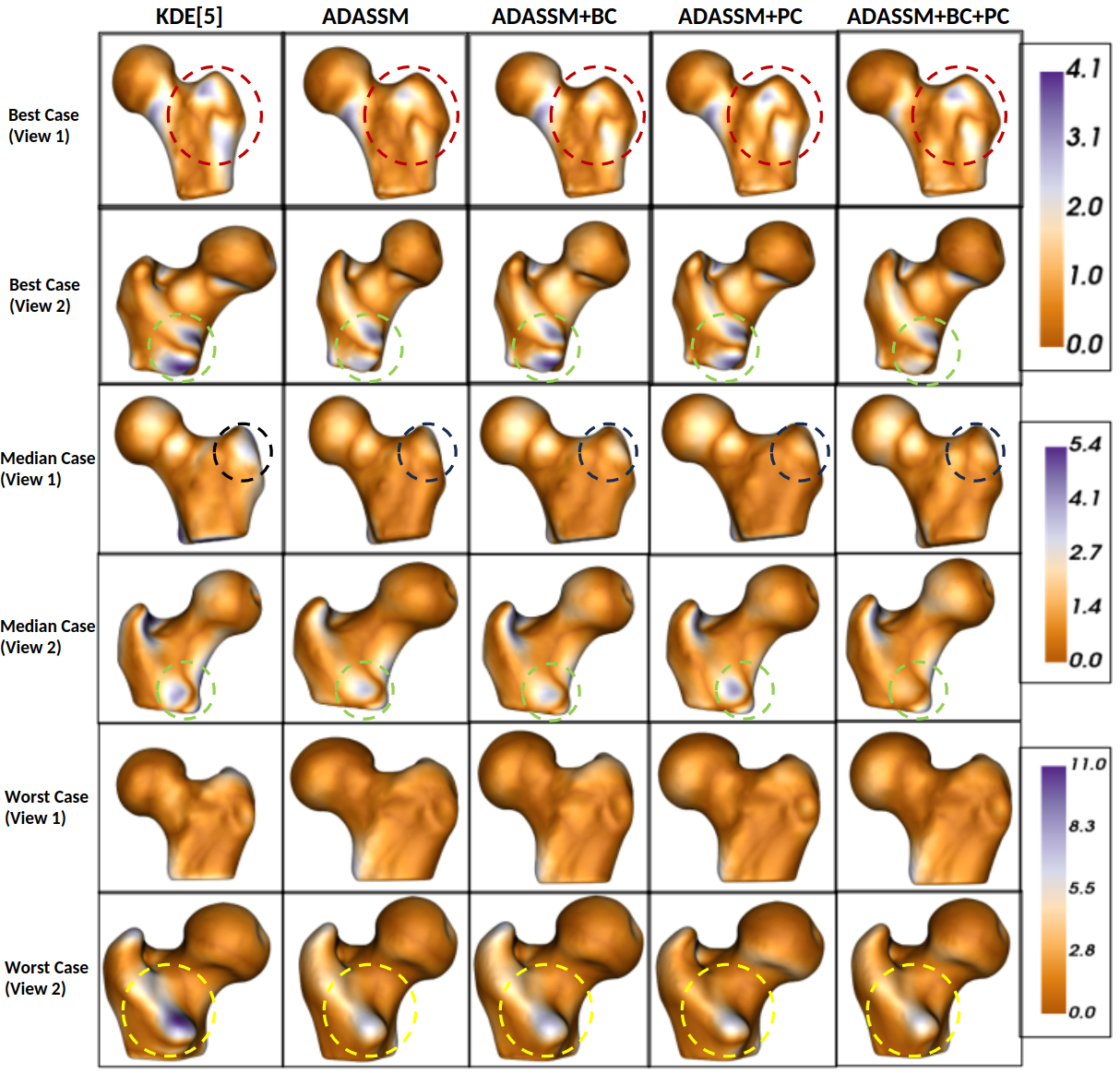}
    \caption{\textbf{Femur Surface-to-Surface distance (mm) visualization:} Shape reconstruction error is displayed as a heatmap on the ground truth reconstructed meshes. The results for different proposed models are presented, showcasing their performance in the best, median, and worst-case scenarios. } 
    \label{fig:result:femur_views}
\end{figure}

Visualizations of surface-to-surface distance for the best, median, and worst cases for the test set are shown in Figure~\ref{fig:result:femur_views}. Upon careful examination of these visualizations, we can observe that the proposed models demonstrate a remarkable reduction in errors in critical regions such as the greater trochanter, growth plate, femoral neck, and epiphyseal lines in the best-case scenario. In the median case, a detailed analysis of both views reveals that in view (1), the KDE baseline \cite{bhalodia2021deepssm} exhibits some errors around the trochanter region that are substantially reduced by the proposed models. In view (2), the error around the trochanter region is completely reduced with the ADASSM+BC+PC model. In the worst-case scenario, view (2) displays the majority of errors in the KDE baseline \cite{bhalodia2021deepssm}, but these errors are significantly diminished, particularly in the lower trochanter region, with the employment of the ADASSM variants.

\begin{figure}[!htb]
   \centering
   \includegraphics[scale=0.27]{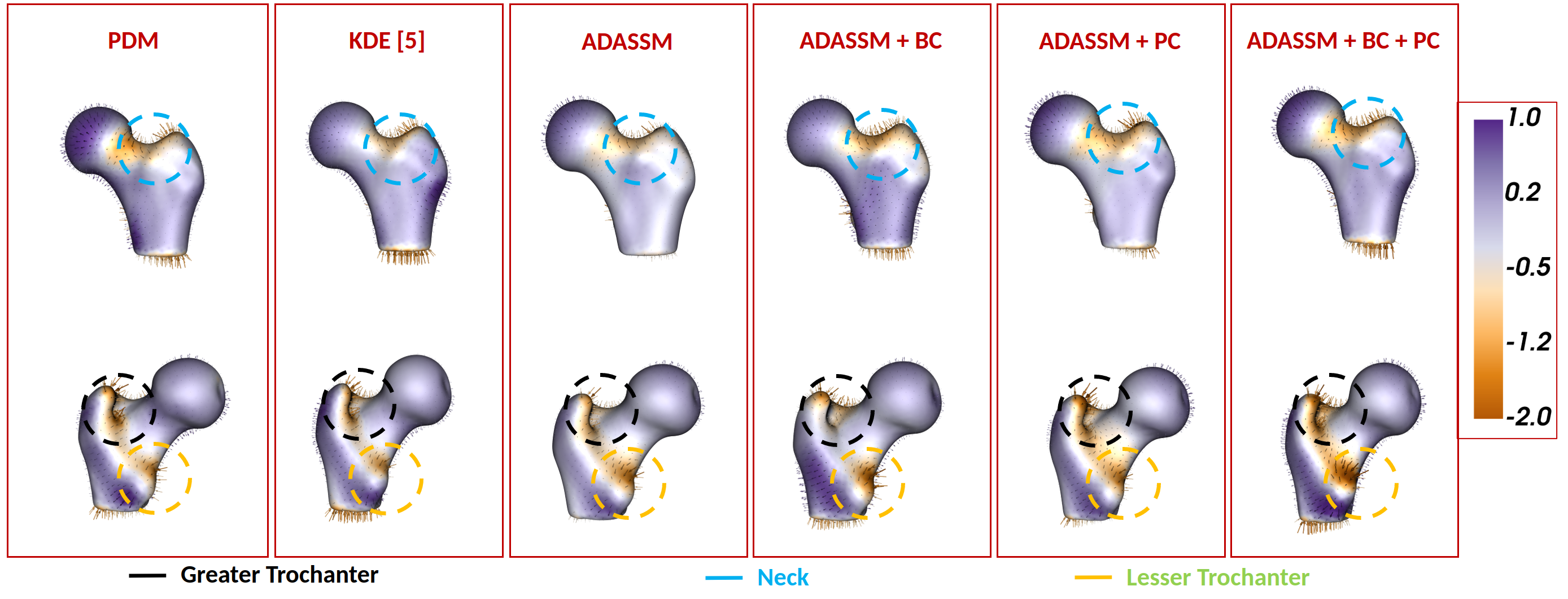}
   \caption{\textbf{Femur shape group difference between CAM-FAI and controls.} The difference $\mu_{cam} - \mu_{normal}$ and is projected on $\mu_{cam}$ with both training and test samples, the arrows denote the direction of the correspondence (particle) movement, and the heatmap showcases the normalized magnitude. }
   \label{fig:result:femur_group_differences}
\end{figure}

\textbf{Downstream Task - Group Differences: }To evaluate the effectiveness of the learned shape representations using the proposed models, we conducted a downstream analysis. The experiments were aimed at evaluating whether the models can accurately capture group differences \cite{harris2013statistical} in medically relevant regions. To achieve this, we formed two groups: a control group and a pathology group consisting of CAM-FAI cases. We calculated the mean differences ($\mu_{normal}$ and $\mu_{cam-FAI}$) between these groups and visualized the differences on a mesh and compared the group differences obtained using the predicted correspondences from the proposed models and the ShapeWorks PDM model. We utilized the entire dataset, including both training and testing samples, for these group differences, and the results are presented in Figure~\ref{fig:result:femur_group_differences}. 

Each group difference illustrates the transition from the mean shape of the pathological group to that of the control group, overlaid on the mean pathological scan. Interestingly, we observed that the group differences between the state-of-the-art PDM model and ADASSM+PC were quite similar compared to baseline KDE \cite{bhalodia2021deepssm}. In some cases, the proposed models exhibited similar differences in medically relevant regions of the femur, whereas in other areas, the models identified additional variations. These findings suggest that the established correspondences can be employed to characterize CAM deformity effectively.

\subsection{Left Atrium}

\hspace{0.15in} \textbf{Data Description and Processing:}
The left atrium dataset consists of 176 late gadolinium enhancement (LGE) MRI images from patients that have been diagnosed with atrial fibrillation (AF). These scans are acquired after the first ablation. 80\% -20\% split is used to split the data where 146 volumes are used to train the model and 30 scans which are used to test the DeepSSM network. To generate point distribution models (or correspondences) for training images, we utilize ShapeWorks \cite{cates2017shapeworks} to optimize a shape model with 1024 correspondences. For training purposes, the MRIs are downsampled from $235 \times 138 \times 175$ (0.625mm isotropic voxel spacing) to $117 \times 69 \times 87$ (1.25mm voxel spacing) by a factor of 2.

\textbf{Training Specifics: }The results for KDE augmentation \cite{bhalodia2021deepssm} on the left atrium dataset are obtained using the default parameters and configuration. With the proposed data augmentation framework, DeepSSM model is trained for 1000 epochs. For the various ADASSM models with the aforementioned regularization losses, a learning rate of 1e-4 is utilized, while ADASSM itself is trained with a learning rate of 5e-3. A batch size of 6 is employed during the training of both the proposed and baseline models. In all ADASSM experiments, $\mathcal{R}$ in the generator is set to 100. Following the publication of the manuscript, we plan to make the training models, implementation code, and relevant hyperparameters publicly available.
 
\begin{figure}[!htb]
    \centering
    \includegraphics[scale=0.39]{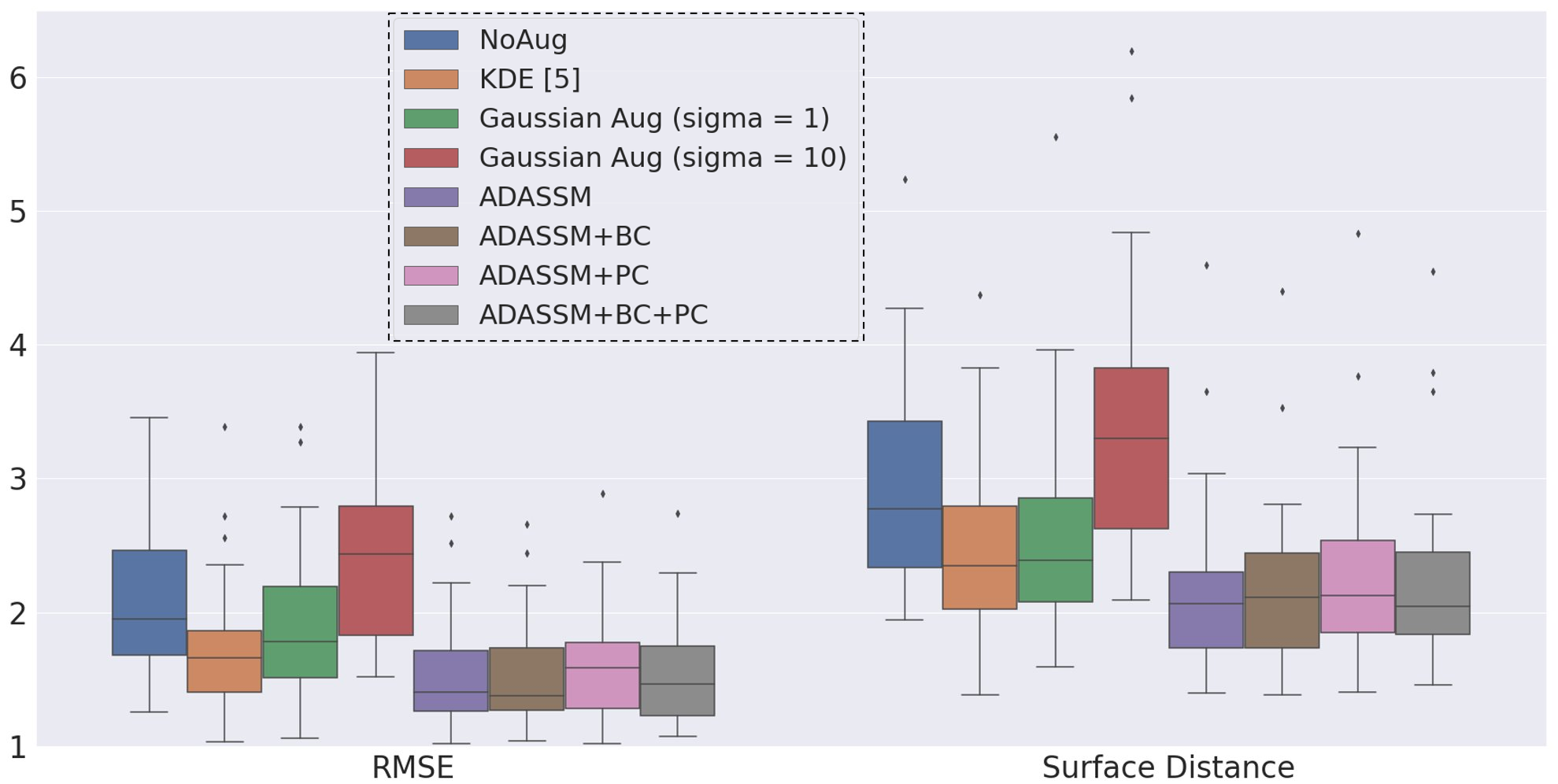}
    \caption{\textbf{Left Atrium Test Set Results}. The \textit{RMSE} is the Euclidean distance between the ground truth and predicted correspondences for all the test samples. The \textit{}{Surface-to-Surface distance (mm)} is computed by comparing the reconstructed mesh using the ground truth correspondences and the predicted correspondences for all the test samples. $\lambda_1$ for ADASSM+BC is 0.001, $\lambda_2$ for ADASSM+PC is 0.05 and $\lambda_1,\lambda_2$ for ADASSM+BC+PC is 0.05. Y-axis is the magnitude of the errors displayed.} 
    \label{fig:result:LA}
\end{figure}

\textbf{Evaluation and Analysis:} Figure~\ref{fig:result:LA} displays the RMSE results alongside the surface-to-surface distance. We can make the following observations from the bar graph:- 1) When augmenting the DeepSSM model with Gaussian noise of varying variances without any adversarial training, we find that it does not improve performance, which may be because the original dataset already has more intensity variation when compared to CT volumes. 2) By enhancing the data- and task-dependency of the noise and integrating the proposed data augmentation framework with various regularization losses, the methodology surpasses the baseline Gaussian noise and KDE shape augmentation framework \cite{bhalodia2021deepssm}. In the left atrium, the performance disparity between the ADASSM variants is more pronounced compared to the femur. This can be attributed to the significant variations observed in the left atrium, where the proposed method excels in effectively regulating the Image-to-SSM task.

\begin{figure}[!htb]
     \centering
    \includegraphics[scale=0.60]{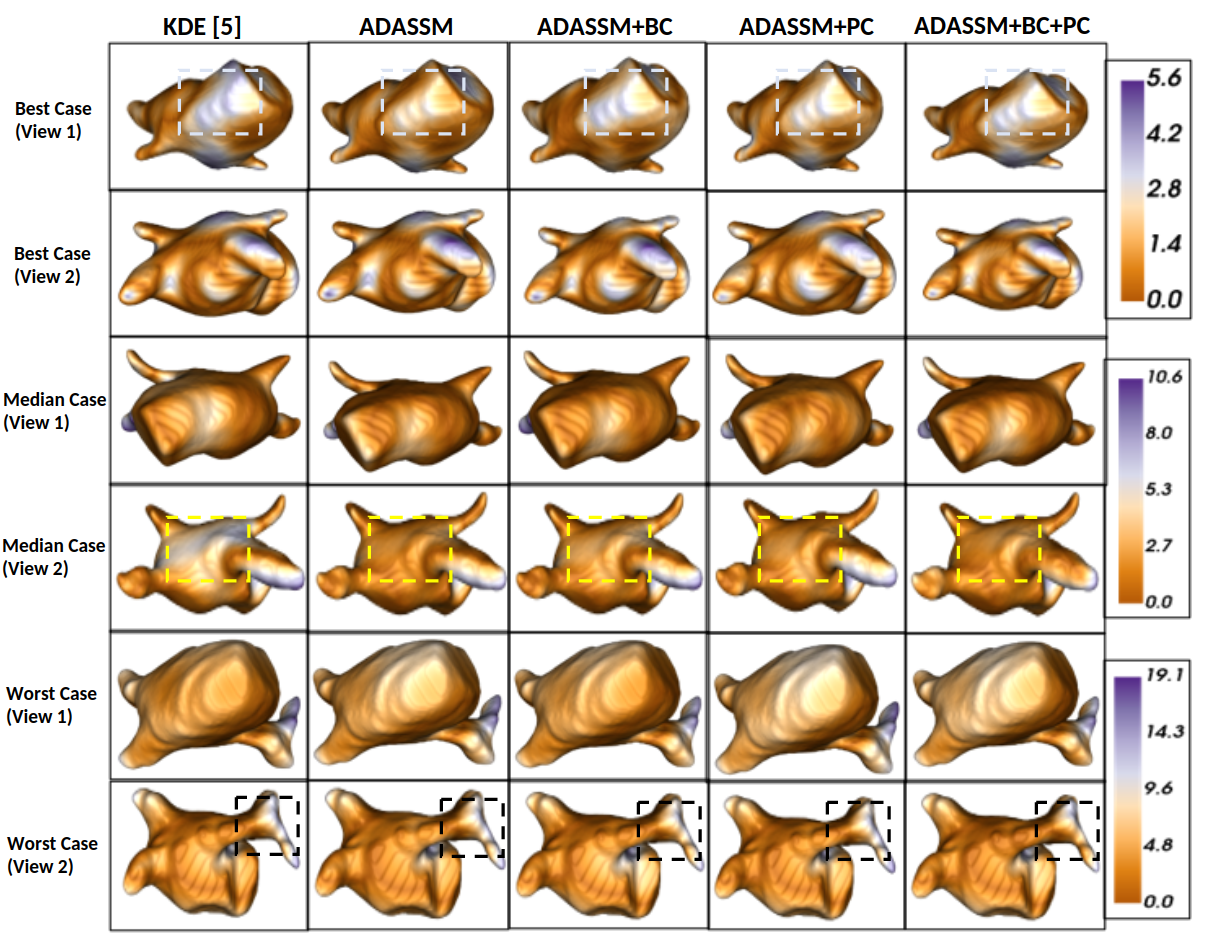}
    \caption{\textbf{Left Atrium Surface-to-Surface distance (mm) visualization:} Shape reconstruction error is displayed as a heatmap on the ground truth reconstructed meshes. The results for the best, median, and worst-case scenarios are shown for various proposed models. We can observe that the proposed methods are outperforming the baseline results in best and median-case.} 
    \label{fig:result:la_best_worst_median}
\end{figure}

In Figure~\ref{fig:result:la_best_worst_median}, visualizations of the surface-to-surface distance are presented for the best, median, and worst cases in the test set. For best-case and median-case views, the proposed model plainly outperforms other methods, while worst-case views are comparable. 

\textbf{Downstream Task - AF Recurrence Prediction: }The shape of the left atrium can provide insights into the recurrence of AF \cite{harris2013statistical}. The dataset has binary outcome labels indicating whether patients experienced AF recurrence after ablation. The goal is to estimate the probability of AF recurrence based on the learned shape representations. We use PCA projections of the shape representations as features for a Multi-Layer Perceptron (MLP) for classification. 
The results are summarized in Table~\ref{tab:1:downstream:LA}. Compared with the traditional SSM \cite{cates2017shapeworks} and KDE, we observe similar performance. All ADASSM variants perform on par with the PDM, with the ADASSM+PC model outperforming the baselines by capturing better shape descriptors for the left atrium than the PCA scores learned in other models. Due to the fact that the classification model is based on the PCA scores of correspondences, ADASSM+PC has the highest accuracy, as the contrastive loss will bring the correspondence's latent space representation closer. But if we train a classifier with non-linear features (other than PCA), ADASSM+BC+PC might result in the best accuracy.

\begin{table}
\centering
\caption{\textbf{AF Recurrence Prediction:} Accuracy of AF recurrence that uses PCA scores as shape descriptors from different models.}
\begin{tabular}{|l|c|}
\hline
\textbf{Model}         &      \textbf{Accuracy}   \\
\hline
ShapeWorks \cite{cates2017shapeworks}       & $53.99\% \pm 6.20$\\
KDE\cite{bhalodia2021deepssm}            & $ 52.66 \% \pm 1.77$\\
Gaussian(sigma=1) & $50.66\% \pm 6.22$ \\
Gaussian(sigma=10)& $ 51.99 \% \pm 7.11$\\
ADASSM          & $51.33\% \pm 2.66$\\
ADASSM+BC       & $55.99\% \pm 15.11$\\
ADASSM+PC       & $67.99\% \pm 2.67$\\
ADASSM+BC+PC    & $ 51.33 \%\pm 2.66$\\
\hline
\end{tabular}
\label{tab:1:downstream:LA}
\end{table}

\subsection{Training Time}

The proposed augmentation method not only improved model performance for both the left atrium and femur datasets but also significantly reduces the training time by approximately 60\% compared to the baseline method \cite{bhalodia2021deepssm} as shown in Table~\ref{tab:2:training:time}. 
\begin{table}
\caption{\textbf{Resources Required:} Comparison for time taken to run \textit{Augmentation} \& \textit{Model} training required on Nvidia RTX 5000 system.}
\centering
   \begin{tabular}{l|c|c}
   %\hline 
    Dataset                  &  KDE[5] & ADASSM \\ \hline
    Left Atrium   & 690.35+122.35 minutes & 336.15 minutes \\ \hline
    Femur  & 725.35+ 93.5 minutes &  399.6 minutes          %\hline
\end{tabular}
\label{tab:2:training:time}
\end{table}
\section{Conclusion and Future Work}
In this study, we introduced a novel methodology by proposing an adversarial data augmentation framework for generic regression tasks with applicability to Image-to-SSM networks. Using data-dependent noise augmentation, the proposed method seeks to discover effective shape representations for three-dimensional volumes. By generating challenging augmentations during model training, the proposed method eliminates the need for offline data augmentation, effectively training a more accurate Image-to-SSM network. The proposed noise augmentation framework outperforms the shape augmentation framework \cite{bhalodia2018deep} and standard noise augmentation, demonstrating that data-dependent noise aids the model by implicitly attending to shape. Through downstream task analysis, we confirmed that the proposed method effectively taught models robust shape descriptors that capture pertinent pathology information. In addition, compared to existing frameworks for shape augmentation, the proposed methodology is not only more robust but also faster. The limitation of the proposed framework is that it trains only on data-dependent intensity/noise augmentations and does not take shape augmentation into account. We plan to extend this framework to data-dependent shape augmentation as well.

\section*{Acknowledgements}
We thank all research members of Dr.Elhabian's lab and the ShapeWorks team for their assistance in discussions and suggestions that helped us improve this work. The National Institutes of Health supported this work under grant numbers NIBIB-U24EB029011, NIAMS-R01AR076120, and NIBIB-R01EB016701. The content is solely the authors' responsibility and does not necessarily represent the official views of the National Institutes of Health.

\bibliographystyle{splncs04}
\bibliography{samplepaper}
\end{document}